\documentclass[11pt,a4paper]{article}
\usepackage[hyperref]{acl2020}
\usepackage{times}
\usepackage{latexsym}

\usepackage{microtype}

\usepackage{latexsym}
\usepackage{multirow}
\usepackage{subfigure}
\usepackage{CJK} 
\usepackage{amsmath}
\usepackage{caption}
\usepackage{booktabs}
\usepackage{amssymb}
\usepackage{graphicx} 
\usepackage{color}

\aclfinalcopy 

\title{Incorporating Uncertain Segmentation Information into Chinese {NER} for Social Media Text}

\author{Shengbin Jia$^{1, 2}$, 
	Ling Ding$^1$, 
	Xiaojun Chen$^1$,
	Shijia E$^2$,
	Yang Xiang$^1$\\
	$^1$Tongji University, Shanghai, China \\
		\texttt{ \{shengbinjia,dling,xiaojunchen,shxiangyang\}@tongji.edu.cn}\\
	$^2$Tencent, Shanghai, China \\
		\texttt{allene@tencent.com}}

\date{}

\begin{document}
\maketitle

\begin{abstract}
	
Chinese word segmentation is necessary to provide word-level information for Chinese named entity recognition (NER) systems. However, segmentation error propagation is a challenge for Chinese NER while processing colloquial data like social media text. In this paper, we propose a model (UIcwsNN) that specializes in identifying entities from Chinese social media text, especially by leveraging uncertain information of word segmentation. Such ambiguous information contains all the potential segmentation states of a sentence that provides a channel for the model to infer deep word-level characteristics. We propose a trilogy (i.e., Candidate Position Embedding $\Rightarrow $ Position Selective Attention $\Rightarrow $ Adaptive Word Convolution) to encode uncertain word segmentation information and acquire appropriate word-level representation. Experimental results on the social media corpus show that our model alleviates the segmentation error cascading trouble effectively, and achieves a significant performance improvement of 2\% over previous state-of-the-art methods.

\end{abstract}

\section{Introduction}

Named entity recognition (NER) is a fundamental task for natural language processing and fulfills lots of downstream applications, such as semantic understanding of social media contents. 
\begin{CJK*}{UTF8}{gbsn} \end{CJK*}
\begin{figure}
	\centering
	\includegraphics[width=0.98\columnwidth]{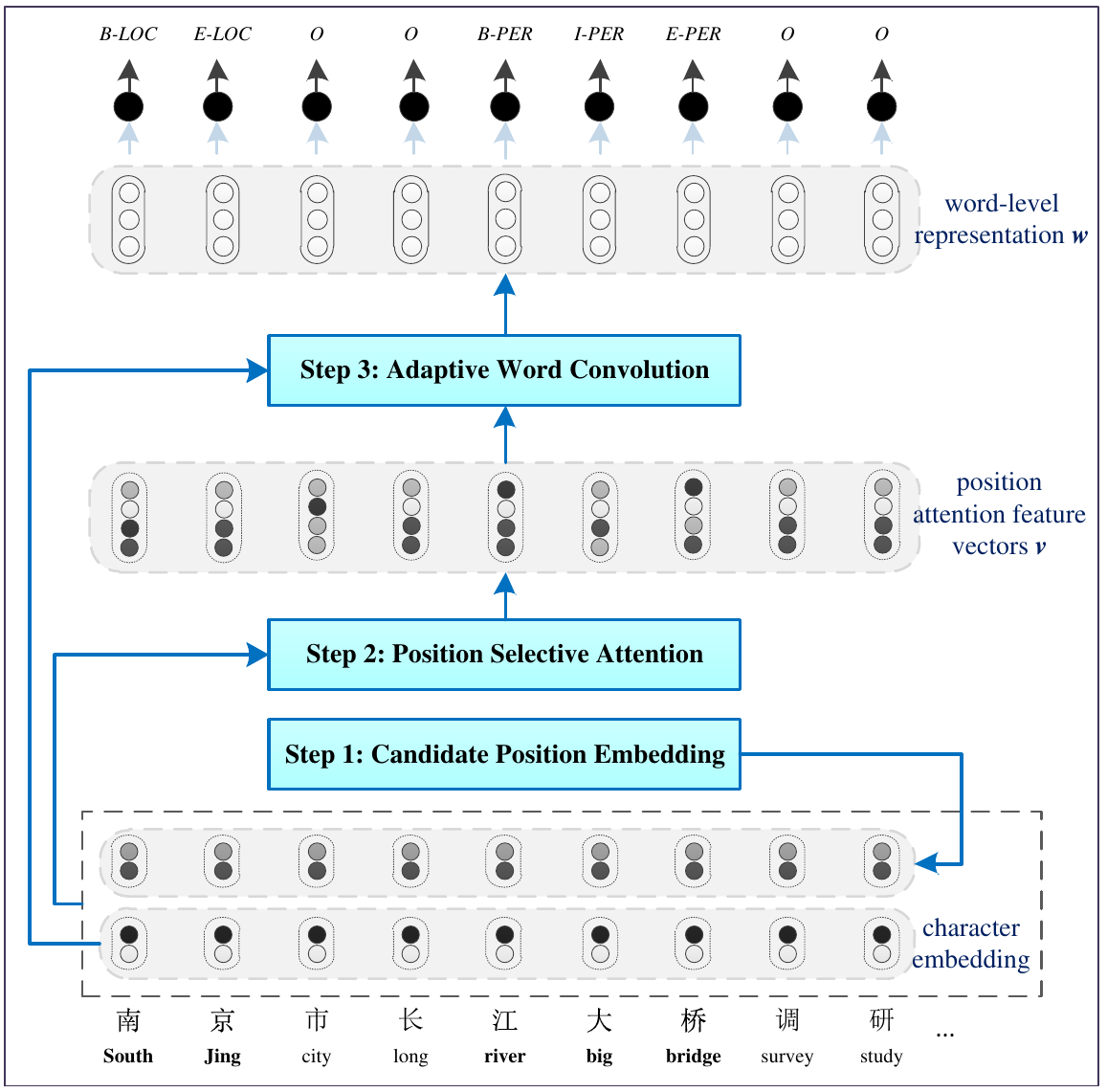}
	\caption{The architecture of our model. An interesting instance ``\begin{CJK*}{UTF8}{gbsn}南京市长江大桥调研\end{CJK*}(Daqiao Jiang, major of Nanjing City, is investigating)..." is represented, which is cited from~\cite{zhang2018chinese}.}
	\label{fig:ChineseNER_3step}
\end{figure}

Chinese NER is often considered as a character-wise sequence labeling task since there are no natural delimiters between Chinese words~\cite{liu2010chinese, li2014comparison}. But the word-level information is necessary for a Chinese NER system~\cite{mao2008chinese,peng2015named,zhang2018chinese}. Various segmentation features can be obtained from the Chinese word segmentation (CWS) procedures then used into a pipeline NER module~\cite{peng2015named, he2017f, zhu2019canner}, or be co-trained by CWS-NER multi-task learning~\cite{peng2016improving, cao2018adversarial}.

However, segmentation error propagation is a challenge for Chinese NER, when processing informal data like social media text~\cite{duan2012cips}. The CWS will produce more unreliable results on the social media text than on the formal data. Incorrectly segmented entity boundaries may lead to NER errors. Nevertheless, most existing extractors always assume that input segmentation information is affirmative and reliable without conscious error discrimination. That is, they acquiesce in that ``The one supposed-reliable word segmentation output of a CWS module will be input into the NER module''. Although the joint training way may improve the accuracy of word segmentations, the NER module still cannot recognize inevitable segmentation errors.

To solve this problem, we design a model (UIcwsNN) that dedicates to identifying entities from Chinese social media text, by incorporating \textbf{U}ncertain \textbf{I}nformation of \textbf{C}hinese \textbf{W}ord \textbf{S}egmentation into a \textbf{N}eural \textbf{N}etwork. 
This kind of uncertain information reflects all the potential segmentation states of a sentence, not just the certain one that is supposed-reliable by the CWS module.  
Furthermore, we propose a trilogy to encode uncertain word segmentation information and acquire word-level representation, as shown in Figure~\ref{fig:ChineseNER_3step}. 

In summary, the contributions of this paper are as follows:

\begin{itemize}

\item We embed candidate position information of characters into the model (in Section~\ref{sec3-1}) to express the states of underlying word.  And we design the Position Selective Attention (in Section~\ref{sec3-2}) that enforces the model to focus on the appropriate positions while ignoring unreliable parts. The above operations provide a wealth of resources to allow the model to infer word-level deep characteristics, rather than bluntly impose segmentation information.
	
\item We introduce the Adaptive Word Convolution (in Section~\ref{sec3-3}), it dynamically provides word-level representation for the characters in specific positions, by encoding segmentations of different lengths. Hence our model can grasp useful word-level semantic information and alleviate the interference of segmentation error cascading.
	
\item Experimental results on different datasets show that our model achieves significant performance improvements compared to baselines that use only character information. Especially, our model outperforms the previous state-of-the-art method by 2\% on the social media.

\end{itemize}

\section{Related Work}

The NER on English has achieved promising performance by naturally integrating character information into word representations~\cite{ma2016end, peters2018deep, yang2018design, yadav2019survey, li2020survey}. However, Chinese NER is still underachieving because of the word segmentation problem. Unlike the English language, words in Chinese sentences are not separated by spaces, so that we cannot get Chinese words without pre-processed CWS. In particular, identifying entities on Chinese social media is harder than on other formal text since there is worse segmentation error propagation trouble. Existing methods payed little attention to this issue, and there were few entity recognition methods specifically for Chinese social media text~\cite{peng2015named,he2017f, he2017unified}.

As for the Chinese NER, existing methods could be classified as either word-wise or character-wise. The former one used words as the basic tagging unit~\cite{ji2005improving}. Segmentation errors would be directly and inevitably entered into NER systems. The latter used characters as the basic tokens in the tagging process~\cite{chen2006chinese, mao2008chinese, lu2016multi, dong2016character}. Character-wise methods that outperformed word-wise methods for Chinese NER~\cite{liu2010chinese, li2014comparison}. 

There were two main ways to take word-level information into a character-wise model. One was to employ various segmentation information as feature vectors into a cascaded NER model. Chinese word segmentation was performed first before applying character sequence labeling~\cite{guo2004chinese,mao2008chinese,zhu2019canner}. The pre-processing segmentation features included character positional embedding~\cite{peng2015named,he2017f,he2017unified}, segmentation tags~\cite{zhang2018chinese,zhu2019canner}, word embedding~\cite{peng2015named,liu2019encoding, xiang2017chinese} and so on. The other was to train NER and CWS tasks jointly to incorporate task-shared word boundary information from the CWS into the NER~\cite{xu2013joint,peng2016improving,cao2018adversarial}.
Although co-training might improve the validity of the word segmentation, the NER module still had no specific measures to avoid segmentation errors. The above existing methods suffered the potential issue of error propagation.

A few researchers tried to address the above defect.
Luo and Yang~\shortcite{luo2016empirical} used multiple word segmentation outputs as additional features to a NER model. However, they treated the segmentations equally without error discrimination.
Liu et al.~\shortcite{liu2019encoding} introduced four naive selection strategies to select words from the pre-prepared Lexicon for their model. However, these strategies did not consider the context of a sentence. 
Zhang and Yang~\shortcite{zhang2018chinese} proposed a Lattice LSTM model that used the gated recurrent units to control the contribution of the potential words. However, as shown by Liu et al.~\shortcite{liu2019encoding}, the gate mechanism might cause the model to degenerate into a partial word-based model.
Ding et al.~\shortcite{DingXZLLS19} and Gui et al.~\shortcite{gui2019lexicon} proposed the models with graph neural network based on the information that the gazetteers or lexicons offered. Obtaining large-scale, high-quality lexicons would be costly. They were dedicated to capturing the correct segmentation information but might not alleviate the interference of inappropriate segmentations.

It is worth mentioning that the above methods were not specifically aimed at social media.
We propose a method to learn word-level representation by leveraging uncertain word segmentation information while considering the informal expression characteristics of social media text.

\section{Methodology}

Figure~\ref{fig:ChineseNER_3step} illustrates the overall architecture of our model UIcwsNN. Given a sentence $S=\left\{c_{1}, c_{2},\cdots, c_{n} \right\}$ as the sequence of characters, each character will be assigned a pre-prepared tag.

We use a Conditional random fields (CRF) layer to decode tags according to the outputs from the sequence encoder~\cite{lample2016neural, yang2018design}. 

As for the sequence encoding, we use the convolution operation as our basic encoding unit. The colloquial social media text usually does not have normative grammar or syntax and presents semantics in fragmented form, for example, ``\begin{CJK*}{UTF8}{gbsn}有好多好多的话想对你说李巾凡想要瘦瘦瘦成李帆我是想切开云朵的心\end{CJK*}(Have many many words to say to you Jinfan Li wanna thin thin thin to Fan Li I am a heart that want to cut the cloud)''. These properties will destroy the propagation of temporal semantic information that comes with the textual sequence. Therefore, the Convolutional neural network (CNN) is naturally suitable for encoding colloquial text because it specializes in capturing salient local features from a sequence.

More importantly, we use a trilogy to learn the word-level representation by incorporating uncertain information of Chinese text segmentation, as shown in the following details.

\subsection{Step-1: Candidate Position Embedding} 
\label{sec3-1}

We design the candidate position embedding to represent candidate positions of each character in all potential words. It reflects the states of all underlying segmentation in a sentence.

\begin{figure}
	\centering
	\includegraphics[width=0.99\columnwidth]{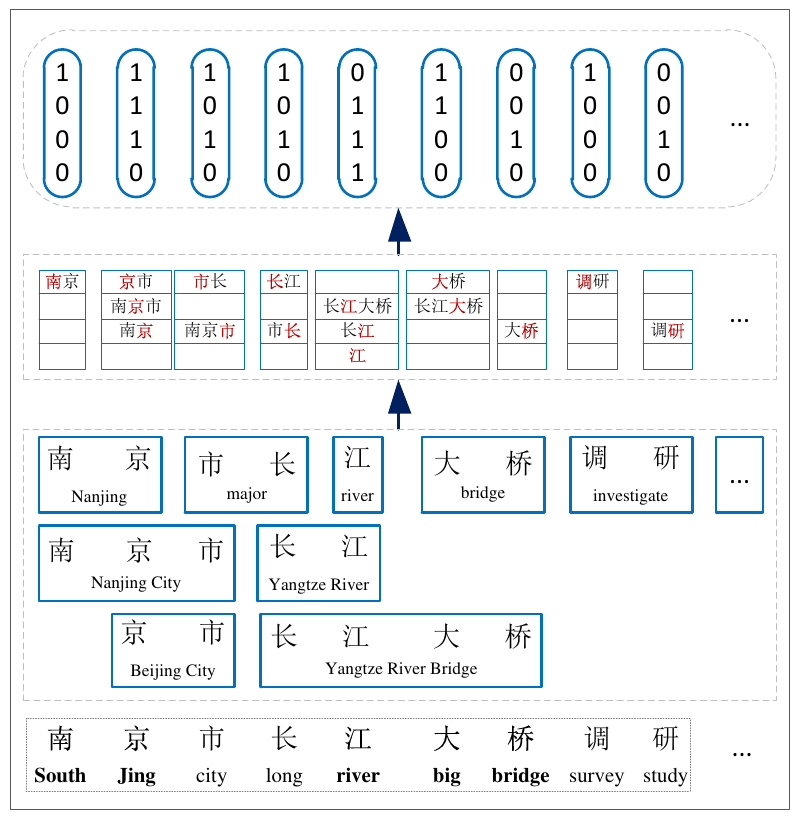}
	\caption{Create the candidate position embedding.}
	\label{fig:ChineseNER_position}
\end{figure}

We firstly scan all the potential words in the sentence that can be worded~\footnote{We use the ``Jieba'', a popular python packages for the CWS. Its special function ``cut\_for\_search()'' can achieve this operation. (https://github.com/fxsjy/jieba)}, so as to obtain as much meaningful segmentation states as possible. As shown in the bottom part of Figure~\ref{fig:ChineseNER_position}, the instance can be segmented and obtained candidate segmentations: ``\begin{CJK*}{UTF8}{gbsn}南京\end{CJK*}(Nanjing), \begin{CJK*}{UTF8}{gbsn}京市\end{CJK*}(Jing City), \begin{CJK*}{UTF8}{gbsn}南京市\end{CJK*}(Nanjing City), \begin{CJK*}{UTF8}{gbsn}市长\end{CJK*}(major), \begin{CJK*}{UTF8}{gbsn}长江\end{CJK*}(Yangtze River), \begin{CJK*}{UTF8}{gbsn}江\end{CJK*}(river), \begin{CJK*}{UTF8}{gbsn}大桥\end{CJK*}(bridge), \begin{CJK*}{UTF8}{gbsn}长江大桥\end{CJK*}(Yangtze River Bridge), \begin{CJK*}{UTF8}{gbsn}调研\end{CJK*}(investigate), ...''.

Next, we use a 4-dimensional vector $\mathbf{c}_{i}^{(p)}$ to embed candidate position information of a character,
where each dimension indicates the positional candidate (i.e., Begin, Inside, End, Single) of a character in words. 1 if it exists, 0 otherwise.
For example, as shown in middle and top parts of Figure~\ref{fig:ChineseNER_position}, as ``\begin{CJK*}{UTF8}{gbsn}京\end{CJK*}(Jing)'' being the begin of ``\begin{CJK*}{UTF8}{gbsn}京市\end{CJK*}(Beijing City)'', the inside of ``\begin{CJK*}{UTF8}{gbsn}南京市\end{CJK*}(Nanjing City)'', and the end of ``\begin{CJK*}{UTF8}{gbsn}南京\end{CJK*}(Nanjing)'', the 1$^{st}$, 2$^{nd}$ and 3$^{rd}$ dimensions of the embedding of ``\begin{CJK*}{UTF8}{gbsn}京\end{CJK*}(Jing)'' are 1, but the 4$^{th}$ dimension is 0 (i.e., $[1, 1, 1, 0]$).

The correct segmentation sequence for the example should be ``\begin{CJK*}{UTF8}{gbsn}南京\end{CJK*}(Nanjing)/\begin{CJK*}{UTF8}{gbsn}市长\end{CJK*}(major)/\begin{CJK*}{UTF8}{gbsn}江大桥\end{CJK*}(Daqiao Jiang)/\begin{CJK*}{UTF8}{gbsn}调研\end{CJK*}(is investigating)/...''. However, the one certain segmentation output that is supposed-reliable by the above CWS tool is ``\begin{CJK*}{UTF8}{gbsn}南京市\end{CJK*}(Nanjing City)/\begin{CJK*}{UTF8}{gbsn}长江大桥\end{CJK*}(Yangtze River Bridge)/\begin{CJK*}{UTF8}{gbsn}调研(investigates)/\end{CJK*}...''. 
The errors may cause that the entity ``\begin{CJK*}{UTF8}{gbsn}江大桥\end{CJK*}(Daqiao Jiang)'' is not recognized.
In contrast, the candidate position embedding should be a more reasonable representation for the Chinese sentence segmentation. It is flexible for a model to infer word-level characteristics.

\subsection{Step-2: Position Selective Attention} 
\label{sec3-2}

There should be only one certain position for a character in the given sentence. We design the position selective attention over candidate positions. It enforces the model to focus on the most relevant positions while ignoring unreliable parts.

Each sequence $S$ is projected to an attention matrix $\boldsymbol{A}$ that captures the semantics of position features interaction according to the contexts. 
\begin{equation}
\boldsymbol{A}=\tanh ( \boldsymbol{W}^{(a)} [\boldsymbol{h}_{1}, \boldsymbol{h}_{2}, \cdots,\boldsymbol{h}_{n}]),
\end{equation}
where $\boldsymbol{A}$ is a matrix of $n\times4$, $\boldsymbol{W}$ is trainable parameters. 

We apply a set of convolution operations that involve filters $\boldsymbol{W}^{(c)} $ and bias terms ${b}^{(c)}$ to the sequence to learn a representation $\boldsymbol{h}_{i}$ for character ${c}_{i}$.
\begin{equation}
\centering
\boldsymbol{h}_{i}=[\boldsymbol{h}_{i}^{l=2}; \boldsymbol{h}_{i}^{l=3}; \boldsymbol{h}_{i}^{l=4}; \boldsymbol{h}_{i}^{l=5}],
\end{equation}
\begin{equation}
\boldsymbol{h}_{i}^{l}=\operatorname{relu}(\boldsymbol{W}_{l}^{(c)}[\boldsymbol{x}_{i}, \cdots, \boldsymbol{x}_{i+l-1}] +{b}_{l}^{(c)}),
\label{euq222}
\end{equation}
where $\mathbf{h}_{i}^{l}$ represents a feature that is generated from a window of length $l$ started with $c_{i}$. The $\boldsymbol{x}_{i}$ is the combination of character embedding $\boldsymbol{c}_{i}^{(e)}$ and expanded candidate position embedding, as
\begin{equation}
\boldsymbol{x}_{i}= \boldsymbol{c}_{i}^{(e)} + \boldsymbol{W}^{(p)} \boldsymbol{c}_{i}^{(p)},
\end{equation}
where $\boldsymbol{c}_{i}^{(e)} \in \mathbb{R}^{d_{e}}$, $\boldsymbol{W}^{(p)} \in \mathbb{R}^{4d_{p}}$.
To enhance the learning of the position information assisted by the character semantic information, we ensure $d_{e} \leqslant d_{p}$.

Given the matrix $\boldsymbol{A}$, we define 
\begin{equation}
\boldsymbol{v}_{i}=\frac{\exp (\boldsymbol{A}_{i,j})}{\sum_{j=0}^{3} \exp (\boldsymbol{A}_{i,j})},
\end{equation}
to quantify the reliability of the $j^{th}$ position with respect to the $i^{th}$ character.

The position attention feature vectors $\boldsymbol{v}$ should assign higher attention values to the appropriate positions while minimizing the values of disturbing positions.

\subsection{Step-3: Adaptive Word Convolution}
\label{sec3-3}

Based on the position selection of each character, the step-3 encodes word segmentations to obtain complete word-level semantics.

As for each character $c_{i}$, we expect to encode the segmentation that involves the $c_{i}$ as its word-level representation. There is a challenge: The lengths of word segmentations are diverse, and the positions of characters located in segmentations are flexible. A single encoding structure is difficult to adapt to this situation. Therefore, we propose the adaptive word convolution.

\begin{figure}
	\centering
	\includegraphics[width=0.47\textwidth]{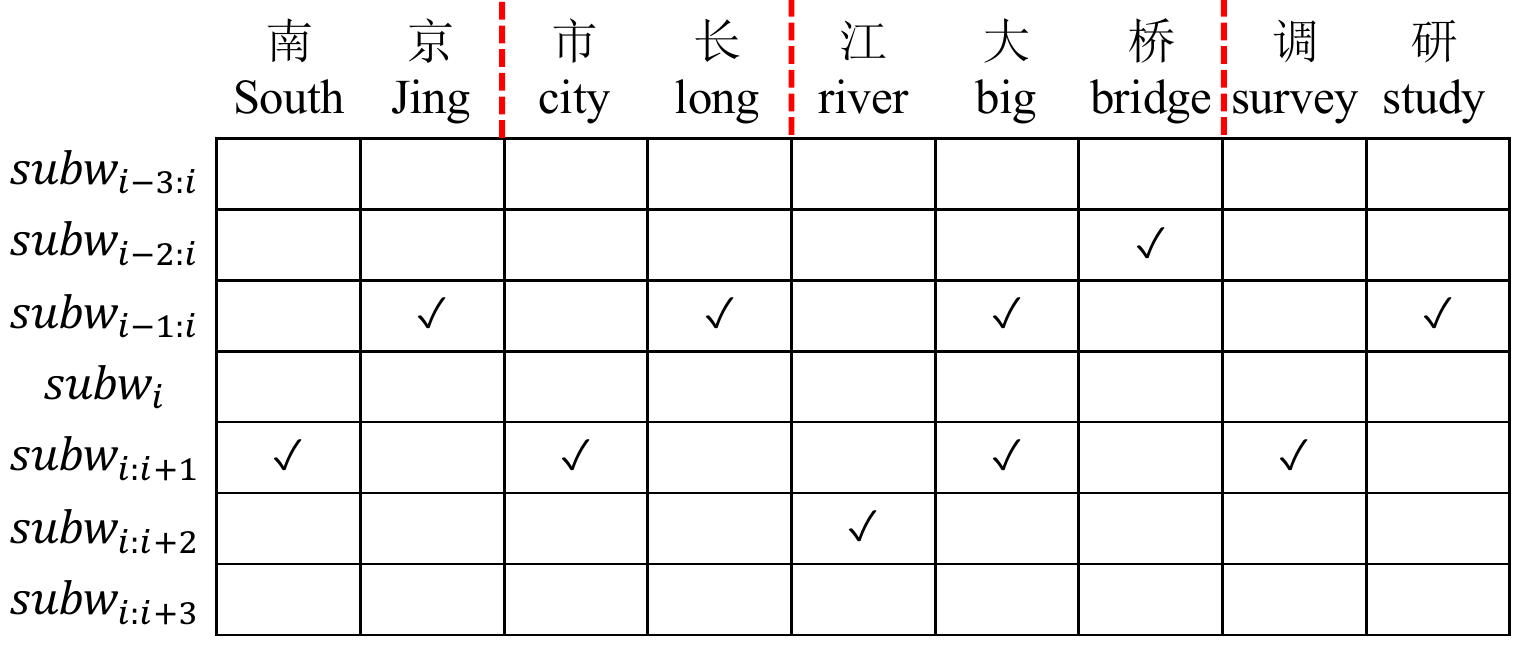}
	\caption{Display the tabulation of subwords. The red vertical lines identify correct word segmentations. The $\checkmark$
shows the subwords that fit each character.}
	\label{fig:subw} 
\end{figure}

When $c_{i}$ is the $k^{th}$ character of the word $w$, we design the word to consist of two parts, namely, the left subword and the right subword, in the form
\begin{equation}
\begin{matrix}
w_{m:m+h-1} \\\\\Leftrightarrow subw_{m:i} \oplus subw_{i:m+h-1} 
\\\\
\Leftrightarrow subw_{(i-k):i} \oplus subw_{i:(i+h-1-k)}, 
\end{matrix}
\end{equation}
where $1 \leqslant m \leqslant n$, $1 \leqslant h \leqslant 4$, \footnote{In most cases, Chinese words are no longer than 4 characters.} $m \leqslant i \leqslant m+h$, and $0 \leqslant k < h$, $\oplus$ denotes join operation.
For the instance mentioned above, we expect to get the tabulation, as shown in Figure~\ref{fig:subw}. For example, the ``\begin{CJK*}{UTF8}{gbsn}南(South)\end{CJK*}'' is the first (i.e., $k=0$) character of the word ``\begin{CJK*}{UTF8}{gbsn}南京\end{CJK*}''(Nanjing) (i.e. $i=m=1$ and $h=2$), we can use the left $subw_{1:1}$ and the right $subw_{1:2}$ to express the word $w_{1:2}$, and then as the word-level representation for the character ``\begin{CJK*}{UTF8}{gbsn}南(South)\end{CJK*}''.
Especially, we discard the $subw_{1:1}$ beacuse $subw_{1:2}$ contains it.

To model subwords automatically, we learn a feature map $\boldsymbol{F}$ ($n\times 7$) through a set of convolution operations with windows of different directions and different sizes, as
\begin{equation}
\boldsymbol{F}=\begin{bmatrix}
	\overleftarrow{\boldsymbol{sw}_{1}^{3}} &  \overleftarrow{\boldsymbol{sw}_{2}^{3}} & \cdots & \overleftarrow{\boldsymbol{sw}_{n}^{3}} \\ 
	\overleftarrow{\boldsymbol{sw}_{1}^{2}} & \overleftarrow{\boldsymbol{sw}_{2}^{2}} &  \cdots & \overleftarrow{\boldsymbol{sw}_{n}^{2}}\\ 
	\overleftarrow{\boldsymbol{sw}_{1}^{1}} & \overleftarrow{\boldsymbol{sw}_{2}^{1}} & \cdots & \overleftarrow{\boldsymbol{sw}_{n}^{1}}\\
	{\boldsymbol{sw}_{1}^{0}} & {\boldsymbol{sw}_{2}^{0}} & \cdots & {\boldsymbol{sw}_{n}^{0}} \\ 
	\overrightarrow{\boldsymbol{sw}_{1}^{1}} & \overrightarrow{\boldsymbol{sw}_{2}^{1}} & \cdots & \overrightarrow{\boldsymbol{sw}_{n}^{1}} \\ 
	\overrightarrow{\boldsymbol{sw}_{1}^{2}} & \overrightarrow{\boldsymbol{sw}_{2}^{2}} & \cdots & \overrightarrow{\boldsymbol{sw}_{n}^{2}}\\ 
	\overrightarrow{\boldsymbol{sw}_{1}^{3}} & \overrightarrow{\boldsymbol{sw}_{2}^{3}} & \cdots & \overrightarrow{\boldsymbol{sw}_{n}^{3}} 
\end{bmatrix},
\end{equation}
\begin{equation}
\overleftarrow{\boldsymbol{sw}_{i}^{k}}=\operatorname{relu}(\boldsymbol{W}_{k}^{(s)}[\boldsymbol{z}_{i-k},\cdots,\boldsymbol{z}_{i}]+{b}_{k}^{(s)}),
\end{equation}
\begin{equation}
\overrightarrow{\boldsymbol{sw}_{i}^{k}}=\operatorname{relu}(\boldsymbol{W}_{k}^{(s')}[\boldsymbol{z}_{i},\cdots,\boldsymbol{z}_{i+k}]+{b}_{k}^{(s')}),
\end{equation}
\begin{equation}
\boldsymbol{z}_{i}= \boldsymbol{c}_{i}^{(e)} + \boldsymbol{W}^{(v)} \boldsymbol{v}_{i},
\end{equation}
where $\boldsymbol{W}^{(v)} \in \mathbb{R}^{d_{v}}$, the $\rightarrow$ indicates the windows sliding forward, whereas $\leftarrow$ shows the windows sliding backward.

Based on the candidate position distribution of characters learned from the step-2, our model can adaptively separate valid subwords from the $\boldsymbol{F}$ to learn the word-level representation $\boldsymbol{w}_{i}$, in detail,

\begin{equation}
\boldsymbol{w}_{i}=\sum_{f=0}^{6} \alpha_{if} \boldsymbol{F}_{i,f},
\end{equation}
\begin{equation}
\alpha_{if}=\frac{\exp (\operatorname{g}(\boldsymbol{F}_{i,f},  \boldsymbol{v}_{i}))}{\sum_{f=0}^{6} \exp (\operatorname{g}(\boldsymbol{F}_{i,f}, \boldsymbol{v}_{i}))},
\end{equation}
\begin{equation}
\operatorname{g}(\boldsymbol{F}_{i}, \boldsymbol{v}_{i})= \tanh ( \boldsymbol{W}^{(\alpha)} [\boldsymbol{F}_{i} + \boldsymbol{W}^{(v)} \boldsymbol{v}_{i}]). 
\end{equation}

After performing the trilogy, the model can grasp useful word-level semantic information and avoid the trouble of segmentation error cascading.

\section{Experiments}

\subsection{Settings}

\textbf{Datasets.} We evaluate Chinese NER models on two popular datasets. The \textit{\textbf{WeiboNER}} corpus~\cite{peng2015named,he2017f}, is drawn from Chinese social media. It contains 1,890 Sina Weibo messages annotated with four entity types ([PER]SON, [ORG]ANIZATION, [LOC]ATION, and [GEO]POLITICAL), including named entities (NAM) and nominal mentions (NOM). The \textbf{\textit{MSRA}} dataset~\cite{levow2006third}, is in the formal text domain. There are 50,729 annotated sentences with three entity types (PER, ORG, and LOC). We use the BIOES scheme (Begin, Inside, Outside, End, Single) to indicate the position of the token in an entity~\cite{Ratinov2009CoNLL}.

\textbf{Evaluation.} We measure the performance of models by regarding three complementary metrics, Precision (P), Recall (R), and F1-measure (F). Each experiment will be performed five times under different random seeds to reduce the volatility of models. Then we report the mean and standard deviation for each model.

\textbf{Hyperparameters.} The character embedding is pre-trained on the raw microblog text~\footnote{http://www.nlpir.org/download/weibo.7z} by the word2vec~\footnote{https://radimrehurek.com/gensim/models/word2vec.html}, and its dimension is 100. As for the base model BiLSTM+CRF, we use hidden state size as 200 for a bidirectional LSTM. As for the base model CNNs+CRF, we use 100 filters with window length \{2, 3, 4, 5\}. We tune other parameters and set the learning rate as 0.001, dropout rate as 0.5. 
We randomly select 20\% of the training set as a validation set. We train each model for a maximum of 120 epochs using Adam optimizer and stop training if the validation loss does not decrease for 20 consecutive epochs.
Besides, we set $d_{e}=d_{p}=100$ and $d_{v}= 25$. We also experiment with other settings and find that these are the most reasonable.

\begin{table*}[]
	\centering
	\renewcommand\arraystretch{1.1}
	\caption{Results of ablation experiments on the WeiboNER dataset and MSRA dataset. The base model is the CNNs+CRF.}
	\resizebox{1.9\columnwidth}{!}{%
	\begin{tabular}{l|lll|lll}
		\toprule[1pt] 
		\multicolumn{1}{c|}{\multirow{2}{*}{Models}} & \multicolumn{3}{c|}{WeiboNER} & \multicolumn{3}{c}{MSRA} \\ \cline{2-7} 
		\multicolumn{1}{c|}{} & P       & R       & F \small{$\pm$std}     & P        & R        & F  \small{$\pm$std}     \\ \hline
		character embedding (baseline)  & 66.45 & 53.47 & 59.22 \small{$\pm$0.42}     & 87.11 & 85.84 & 86.47 \small{$\pm$0.21}\\ \cline{1-7} 
		+ certain segmentation feature (CS) & 68.41 & 51.82 & 58.92 \small{$\pm$0.54}  & 90.37 & 88.06 & 89.20 \small{$\pm$0.12} \\ \cline{1-7} 
		+ candidate position embedding (CPE) & 65.19 & 56.46 & 60.51 \small{$\pm$0.37}  & 90.20 & 88.27 & 89.22 \small{$\pm$0.06}\\ 
		+ position selective attention (PSA) & 68.50 & 55.31 & 61.13 \small{$\pm$0.49} & 90.34 & 89.08 & 89.71 \small{$\pm$0.22}\\
		+ adaptive word convolution (AWC) &67.37 &57.61 &62.07 \small{$\pm$0.61} & 89.87 & 90.54 & 90.20 \small{$\pm$0.24}\\ \cline{1-7} 
		base model + BERT & 78.01 & 72.97 & 75.40 \small{$\pm$0.33}& 94.51 & 91.72 & 93.09 \small{$\pm$0.27} \\ 
		UIcwsNN + BERT & 79.64 & 73.29 & 76.33 \small{$\pm$0.20} & 96.31 & 94.98 & 95.64 \small{$\pm$0.15}\\
		\bottomrule[1pt]
	\end{tabular}
	}
	\label{table:1}
\end{table*}

\subsection{Results and Detailed Analysis}

\subsubsection{Ablation Study}

To study the contribution of each component in our model, we conduct ablation experiments on the two datasets where we use the product of each step to decode tags. We display the results in Table~\ref{table:1} and draw the following conclusions.

The feature (CS) is generated from the one certain segmentation output that is supposed-reliable by the CWS tool Jieba, and it may not benefit the NER on social media text. Compared with the corresponding baseline, the feature (CS) impels the model to improve its performance on the MSRA dataset but to reduce performance on the WeiboNER corpus. There are more segmentation errors on social media text than on formal text so that the impact of error cascading is heavy for NER on social media.

On the WeiboNER dataset, the three steps exert different capabilities for improving model performance.
Compared with the baseline, the model with the step-1 (+CPE) yields 1.3\% improvement in the F value, and its recall improves significantly by 3\%, although the precision decreases 1.2\%.
After we continue with the step-2 (+ PSA), the F value further increases by 0.6\%. In this scenario, both precision and recall are higher than the baseline.
When the step-3 (+AWC) is completed, the F value further increases by 0.9\%. In this scenario, the recall significantly improves by 4\% with 0.9\% improvement in precision, compared to the baseline.

Combining the results on the two different datasets, we find several consistent phenomena. Globally, the F values of the model keep increasing after each step. From a decomposition perspective, the step-2 (+PSA) is notable for improving the precision of the model. And the step-3 (+AWC) is significant for improving the recall. Therefore, the trilogy is complementary.

Our method has good robustness. On the two datasets from different domains, the uncertain information of word segmentations is always efficient, the trilogy (i.e., +CPE, +PSA, +AWC) is valuable. However, performance improvement on the WeiboNER dataset is more significant than on the MSRA dataset. In contrast with formal text, the social media text contains more word segmentation errors that better reflects the advantages of our method.

Finally, We verify the influence of the pre-trained language model BERT~\cite{devlin2018bert} on our model.
We optimize the BERT \footnote{https://storage.googleapis.com/bert\_models/2018\_11\_03/ chinese\_L-12\_H-768\_A-12.zip}
to obtain the character embedding and train the model CNNs+CRF jointly, where its F value reaches 75\% on the WeiboNER dataset. The BERT improves the entity recognition outcome dramatically since it uses large-scale external data to pre-train the contextual embedding.
When we use our model UIcwsNN to replace the base model CNNs+CRF, the effect is improved by nearly 1\%.
It proves that our trilogy and the BERT are complementary. The BERT can provide high-quality character-level embedding to the model, and our method contributes word-level semantic information for the model.
This conclusion can also be drawn from the results of the MSRA dataset.

\begin{table}
	\centering
	\renewcommand\arraystretch{1.1}
	\caption{The F values of existing models on the WeiboNER dataset. $^{*}$ indicates that the model utilizes external lexicons. $^{\circ}$ indicates that the model adopts joint learning. The previous models do not use the BERT, so we show the results of our model without BERT.}
	\resizebox{0.995\columnwidth}{!}{%
		\begin{tabular}{lc|ccc}
			\toprule[1pt] 
			\multicolumn{2}{l|}{Models} & NAM & NOM     & Overall     \\ \hline
			\multicolumn{2}{l|}{\cite{peng2015named}$^{\circ}$} &  51.96     &   61.05   & 56.05    \\\hline
			\multicolumn{2}{l|}{\cite{peng2016improving}$^{\circ}$}&  55.28     &   62.97    &  58.99     \\ \hline
			\multicolumn{2}{l|}{\cite{he2017f}}&   50.60    &   59.32    &  54.82    \\ \hline
			\multicolumn{2}{l|}{\cite{he2017unified}} &  54.50     &   62.17    &  58.23     \\ \hline
			\multicolumn{2}{l|}{\cite{zhang2018chinese}$^{*}$} &  53.04     &   62.25    &   58.79    \\ \hline
			\multicolumn{2}{l|}{\cite{cao2018adversarial}$^{\circ}$}   &   54.34    &   57.35    &    58.70   \\ \hline
			\multicolumn{2}{l|}{\cite{zhu2019canner}}&   55.38   &  62.98    &    59.31   \\ \hline
			\multicolumn{2}{l|}{\cite{liu2019encoding}$^{*}$} &   52.55    &   \textbf{67.41}    &    59.84   \\ \hline
			\multicolumn{2}{l|}{\cite{DingXZLLS19}$^{*}$} & -& -& 59.50\\ \hline
			\multicolumn{2}{l|}{\cite{gui2019lexicon}$^{*}$} &55.34&64.98& 60.21\\ \hline
			\multicolumn{2}{l|}{\cite{johnson2020cwpcbiatt}} &55.70&62.80& 59.50\\ \hline
			
			BiLSTM+CRF & & 53.95 & 62.63 & 57.69 \\
			CNNs+CRF & & 55.07 & 62.97  & 59.22 \\
			Our model (UIcwsNN) & & \textbf{57.58}	&	\textbf{65.97}  & \textbf{62.07} \\
			\bottomrule[1pt]
		\end{tabular}
	}
	\label{table:2}
\end{table}

\begin{table}
	\centering
	\renewcommand\arraystretch{1.1}
	\caption{The results of different models on the MSRA dataset. $^{\times }$ indicates that the model uses the BERT.}
	\resizebox{0.96\columnwidth}{!}{%
		\begin{tabular}{l|ccc}
			\toprule[1pt] 
			Model & P & R     & F     \\ \hline
			\cite{chen2006chinese} &91.22 &	81.71 &	86.20 \\ \hline
			\cite{dong2016character} &  91.28& 90.62 & 90.95 \\ \hline
			\cite{zhang2018chinese} &  93.57     &   92.79    &   93.18    \\ \hline
			\cite{zhu2019canner} &   93.53 &  92.42   &    92.97   \\ \hline
			\cite{DingXZLLS19} & 94.60 & 94.20 & 94.40\\ \hline
			\cite{8855350}$^{\times}$ &  95.46&	95.09&	95.28   \\ \hline
			\cite{gong2019chinese}$^{\times}$&  95.26 & \textbf{95.57} & 95.42     \\ \hline
			\cite{johnson2020cwpcbiatt} &  93.71 & 92.29 & 92.99\\ \hline
			
			UIcwsNN   & 89.87 & 90.54 & 90.20  \\
			UIcwsNN + BERT$^{\times}$  &  \textbf{96.31} & \textbf{94.98} & \textbf{95.64} \\ 
			\bottomrule[1pt]
		\end{tabular}
	}
	\label{table:3}
\end{table}

\subsubsection{Comparison with Existing Methods}

Table~\ref{table:2} represents the results of the WeiboNER dataset.
Our model UIcwsNN significantly outperforms other models and achieves new state-of-the-art performance. The overall score of our model is generally more than 2\% higher than the scores of other models.
Many methods use lexicon instead of the CWS to provide extractors with external word-level information, but how to choose the appropriate words based on sentence contexts is their challenge. Besides, the approaches that jointly train NER and CWS tasks do not achieve desired results, because segmentation noises affect their effectiveness inevitably. Our model handles this trouble.

The CNN-based models achieve better performance compared to the model BiLSTM+CRF. Furthermore, most of the existing methods construct encoders based on recurrent neural networks or graph neural networks. Although they perform excellent results on the MSRA dataset, they do not achieve a significant improvement on the WeiboNER corpus. In addition to the word segmentation error propagation on social media, another important reason may be that the fragmented semantic expression of colloquial text limits their performance. In contrast, our CNN-based model plays a better advantage in capturing the fragmented semantics of colloquial text.

Results on the MSRA dataset are shown in Table~\ref{table:3}. Our model UIcwsNN specializes in learning word-level representation, but rarely considers other-levels characteristics, such as long-distance temporal semantics. Therefore, it only achieves competitive performance on the formal text. But our model UIcwsNN+BERT realizes new state-of-the-art performance.

\begin{figure}
	\centering
	\includegraphics[width=0.48\textwidth]{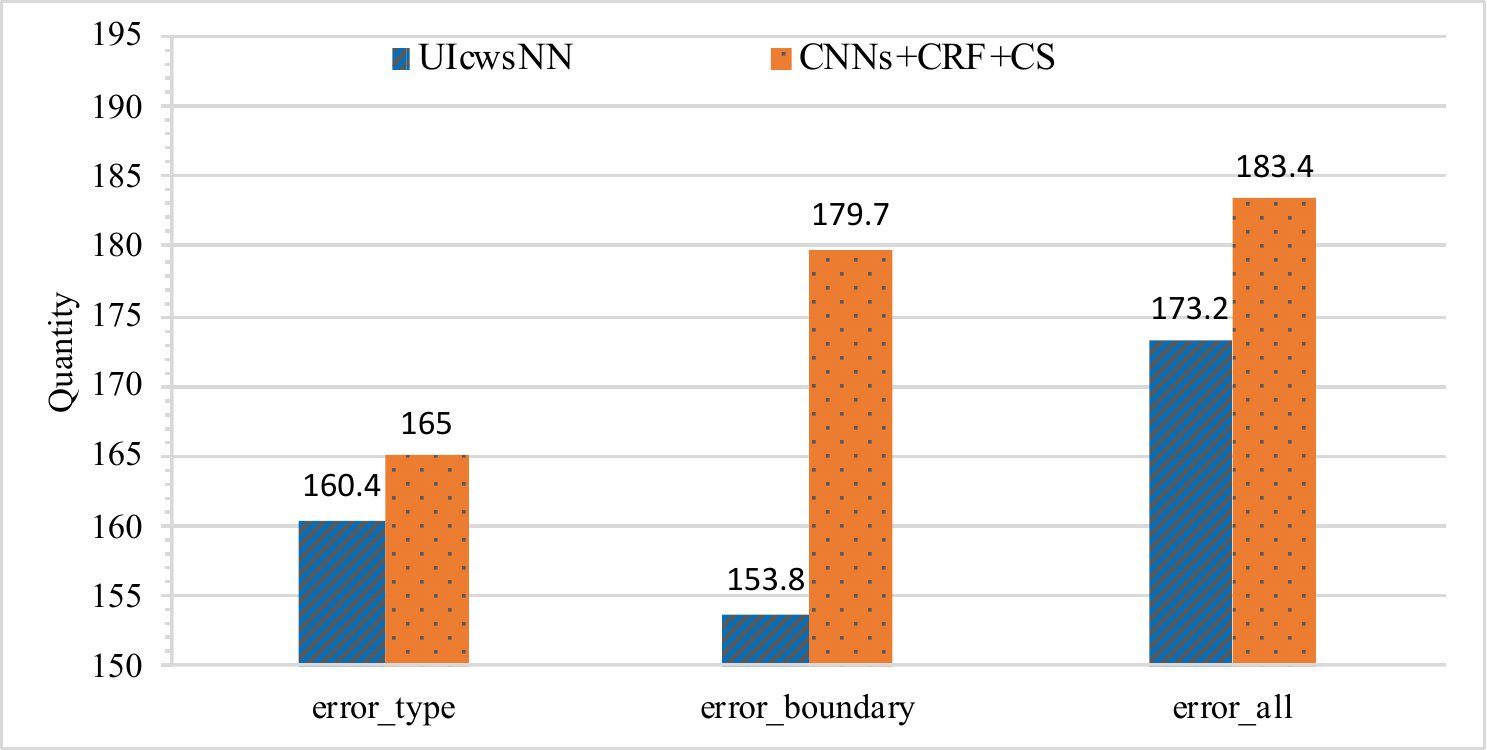}
	\caption{The statistics of the model output errors on the WeiboNER corpus. The model CNNs+CRF+CS uses the feature of the one supposed-reliable word segmentation output from the CWS tool Jieba.}
	\label{fig:two} 
\end{figure}

\subsubsection{Error Analysis}

We count the output errors of models and classify them into two categories~\footnote{If there are two kinds of errors on a predicted entity, the error will be counted twice.}: type error and boundary error, as shown in Figure~\ref{fig:two}. The model CNNs+CRF+CS produces more boundary errors than type errors. However, our model UIcwsNN dramatically decreases the boundary error outputs (and the type errors are also reduced), so that the error distribution is reversed. That is, in model UIcwsNN, the proportion of boundary errors is smaller than that of type errors, but in model CNNs+CRF+CS, the opposite is true. This situation shows that word segmentation errors generated by the word segmentation tool seriously affect model performance, especially misleading the model to identify wrong entity boundaries. Our method can learn the word boundaries effectively, thereby alleviating the cascade of segmentation errors.

\begin{figure}
	\centering
	\includegraphics[width=0.48\textwidth]{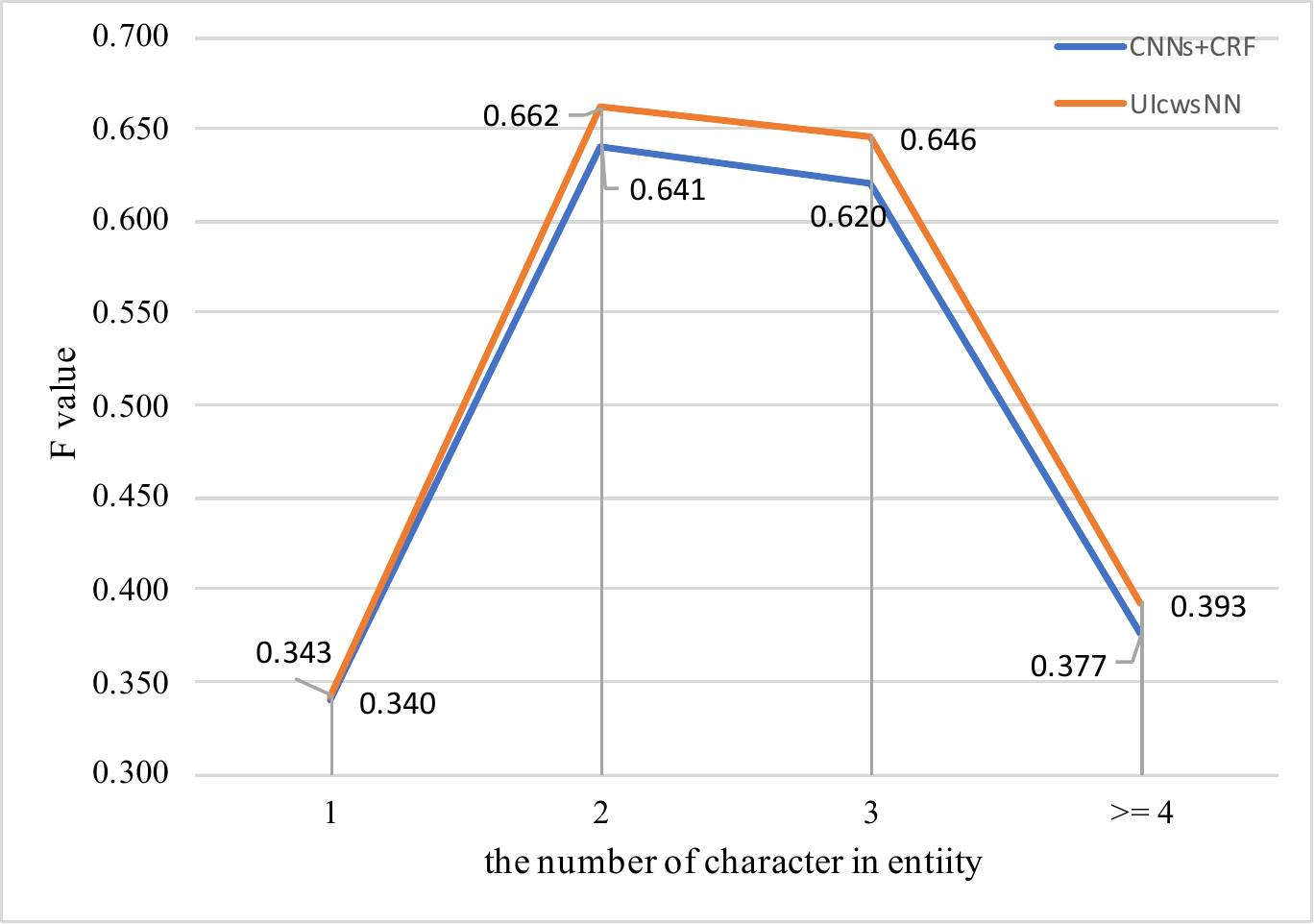}
	\caption{Performance of multi-character entities on the WeiboNER dataset. The base model CNNs+CRF only uses character embedding.}
	\label{fig:multichar}
\end{figure}

\begin{table*}[]
	\centering
	\renewcommand\arraystretch{1.2}
	\caption{Testing examples with segmentation errors.}
	\resizebox{2\columnwidth}{!}{%
		\begin{tabular}{l|ccccccccccccc}
			\toprule[1pt] 
			\multirow{2}{*}{Case One} & \multicolumn{13}{l}{\begin{CJK*}{UTF8}{gbsn}有人祝我早生贵[\textbf{女}]$_{PER.NOM}$真是无语啊\end{CJK*}}  \\ 
			& \multicolumn{13}{l}{Someone wished me to have a precious daughter soon, I am so speechless} \\ \hline
			\multirow{2}{*}{candidate segmentation}  
			& \multicolumn{13}{l}{\begin{CJK*}{UTF8}{gbsn}有人(someone), 祝(wish), 我(me), 早(soon), 早生(early birth), 生贵(precious), 贵(pre-\end{CJK*}} \\ 
			& \multicolumn{13}{l}{\begin{CJK*}{UTF8}{gbsn}cious), \textbf{女}(daughter), \textbf{女}真(Nuzhen), 是(is), 真是(really), 无语(speechless), 啊(ah)\end{CJK*}} 
			\\ \hline
			\multirow{2}{*}{one certain segmentation}
			& \multicolumn{13}{l}{\begin{CJK*}{UTF8}{gbsn}有人(someone), 祝(wish), 我(me), 早(soon), 生贵(precious), \textbf{女}真(Nuzhen), 是(is), 无\end{CJK*} } \\ 
			& \multicolumn{13}{l}{\begin{CJK*}{UTF8}{gbsn}语(speechless), 啊(ah)\end{CJK*} } \\
			\bottomrule[1pt]
			\multirow{2}{*}{Case Two} & \multicolumn{13}{l}{\begin{CJK*}{UTF8}{gbsn}刚刚获得了微博[\textbf{准会员}]$_{PER.NOM}$专属徽章，开心\end{CJK*}}  \\
			& \multicolumn{13}{l}{I just got the exclusive badge for a weibo associate member, I am happy} \\ \hline
			\multirow{2}{*}{candidate segmentation} 
			& \multicolumn{13}{l}{\begin{CJK*}{UTF8}{gbsn}刚刚(just now), 获得(get), 了(finish), 微博(weibo), 微博\textbf{准}(wei bo zhun), \textbf{准会}(quasi),\end{CJK*} } \\
			& \multicolumn{13}{l}{\begin{CJK*}{UTF8}{gbsn}\textbf{会员}(member), 专属(exclusive),  徽章(badge), 开心(happy)\end{CJK*} } \\\hline
			\multirow{2}{*}{one certain segmentation} 
			& \multicolumn{13}{l}{\begin{CJK*}{UTF8}{gbsn}刚刚(just now), 获得(get), 了(finish), 微博\textbf{准}(wei bo zhun), \textbf{会员}(member), 专属(exc-\end{CJK*}} \\
			& \multicolumn{13}{l}{\begin{CJK*}{UTF8}{gbsn}lusive), 徽章(badge), 开心(happy)\end{CJK*}} \\
			\bottomrule[1pt]
		\end{tabular}
	}
	\label{table:42}
\end{table*}

\subsubsection{Performance against Multi-character Entities} 

Figure~\ref{fig:multichar} shows the performance of recognizing entities with different lengths \{1, 2, 3, $\geqslant$4\}. 
According to statistics, entities with two or three characters account for more than 95\% of the total number of entities.
Both models give high F scores for entities of moderate lengths \{2, 3\}, but low performance for entities that are too short or too long. The reasons may be that entities with a single character or more than four characters are rare, resulting in model training inadequately. Our model UIcwsNN achieves better results than the base model CNNs+CRF when identifying entities of various lengths. In particular, as for entities with two or three characters, the model UIcwsNN yields more than 2\% improvement. This situation implies that our model captures word-level semantic information by modeling the uncertain information of word segmentations so that it is good at recognizing multi-character entities.

\subsubsection{Case Study}

\begin{figure}
	\centering
	\includegraphics[width=0.48\textwidth]{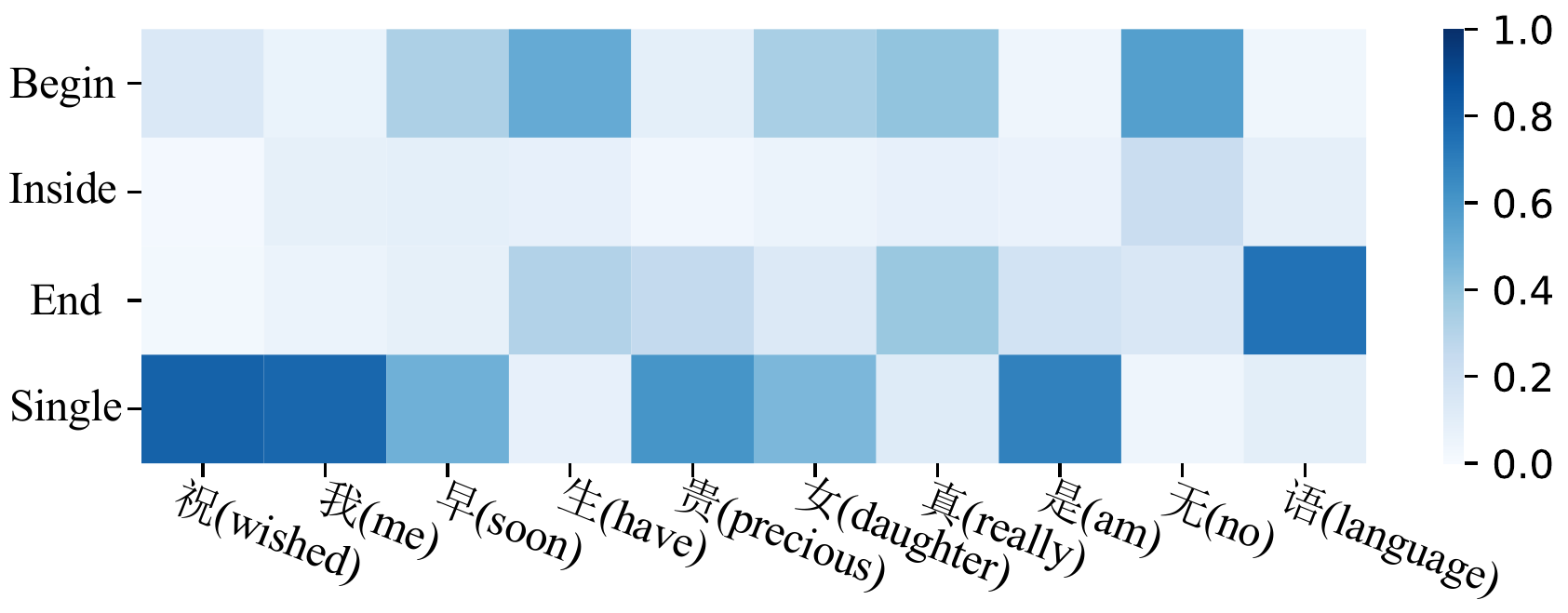}
	\caption{Visualization of position attention values $\boldsymbol{v}$ obtained from the position selective attention.}
	\label{fig:three} 
\end{figure}

Table~\ref{table:42} shows several examples with word segmentation errors.
When we use the one certain (supposed-reliable) segmentation sequence from the tool Jieba as the word-level feature for the model CNNs+CRF+CS,  the segmentation errors ``\begin{CJK*}{UTF8}{gbsn}女真\end{CJK*}'(Nuzhen)" and ``\begin{CJK*}{UTF8}{gbsn}微博准\end{CJK*}(wei bo zhun)'' lead to the misjudgments of the entities ``\begin{CJK*}{UTF8}{gbsn}女\end{CJK*}(daughter)'' and ``\begin{CJK*}{UTF8}{gbsn}准会员\end{CJK*}(associate member)'', respectively. Our model UIcwsNN can extract these entities. The uncertain character positions can provide our model with rich word-level information. Then, we use the position selective attention to support the model to learn appropriate segmentation states.
The visualization of the first case in Figure~\ref{fig:three} shows that our model can assign higher attention values to the appropriate positions while mitigating error interferences.

\section{Conclusion}

Named entity recognition is an urgent task for semantic understanding of social media content. As for the Chinese NER, Chinese word segmentation error propagation is prominent since there is much colloquial text in social media. 
In this paper, we explore a trilogy to leverage the uncertain information of word segmentation to avoid the interference of segmentation errors. The step-1 utilizes the Candidate Position Embedding to present the potential segmentation states of a sentence; The step-2 employs the Position Selective Attention to capture appropriate segmentation states while ignoring unreliable parts; The step-3 uses the Adaptive Word Convolution to encode word-level representation dynamically.
We analyze the performance of each component of the model and discuss the relationship between the model and related factors such as segmentation error, BERT, and entity length. 
Experiment results on different datasets show that our model achieves new state-of-the-art performance. It demonstrates that our method has an excellent ability to capture word-level semantics and can alleviate the segmentation error cascading trouble effectively.
In future work, we hope that the model can get rid of the word segmentation tool, instead, learn the candidate position informationn autonomously. We will release the source code when the paper is openly available.

\section*{Acknowledgments}
This work is supported by the National Key Research and Development Project of China under Grant no. 2019YFB1704402, the 2019 Tencent Marketing Solution Rhino-Bird Focused Research Program, and the 2020 Tencent Rhino-Bird Elite Training Program.

\bibliography{cnerACL2020}

\begin{thebibliography}{34}
\expandafter\ifx\csname natexlab\endcsname\relax\def\natexlab#1{#1}\fi

\bibitem[{Cao et~al.(2018)Cao, Chen, Liu, Zhao, and Liu}]{cao2018adversarial}
Pengfei Cao, Yubo Chen, Kang Liu, Jun Zhao, and Shengping Liu. 2018.
\newblock Adversarial transfer learning for chinese named entity recognition
  with self-attention mechanism.
\newblock In \emph{Proceedings of the 2018 Conference on Empirical Methods in
  Natural Language Processing}, pages 182--192.

\bibitem[{Chen et~al.(2006)Chen, Peng, Shan, and Sun}]{chen2006chinese}
Aitao Chen, Fuchun Peng, Roy Shan, and Gordon Sun. 2006.
\newblock Chinese named entity recognition with conditional probabilistic
  models.
\newblock In \emph{Proceedings of the Fifth SIGHAN Workshop on Chinese Language
  Processing}, pages 173--176.

\bibitem[{Devlin et~al.(2018)Devlin, Chang, Lee, and
  Toutanova}]{devlin2018bert}
Jacob Devlin, Ming-Wei Chang, Kenton Lee, and Kristina Toutanova. 2018.
\newblock Bert: Pre-training of deep bidirectional transformers for language
  understanding.
\newblock \emph{arXiv preprint arXiv:1810.04805}.

\bibitem[{Ding et~al.(2019)Ding, Xie, Zhang, Lu, Li, and Si}]{DingXZLLS19}
Ruixue Ding, Pengjun Xie, Xiaoyan Zhang, Wei Lu, Linlin Li, and Luo Si. 2019.
\newblock \href {https://www.aclweb.org/anthology/P19-1141/} {A neural
  multi-digraph model for chinese {NER} with gazetteers}.
\newblock In \emph{Proceedings of the 57th Conference of the Association for
  Computational Linguistics, {ACL} 2019}, pages 1462--1467.

\bibitem[{Dong et~al.(2016)Dong, Zhang, Zong, Hattori, and
  Di}]{dong2016character}
Chuanhai Dong, Jiajun Zhang, Chengqing Zong, Masanori Hattori, and Hui Di.
  2016.
\newblock Character-based lstm-crf with radical-level features for chinese
  named entity recognition.
\newblock In \emph{Natural Language Understanding and Intelligent
  Applications}, pages 239--250. Springer.

\bibitem[{Duan et~al.(2012)Duan, Sui, Tian, and Li}]{duan2012cips}
Huiming Duan, Zhifang Sui, Ye~Tian, and Wenjie Li. 2012.
\newblock The cips-sighan clp 2012 chineseword segmentation onmicroblog corpora
  bakeoff.
\newblock In \emph{Proceedings of the second CIPS-SIGHAN joint conference on
  Chinese language processing}, pages 35--40.

\bibitem[{E and Xiang(2017)}]{xiang2017chinese}
Shijia E and Yang Xiang. 2017.
\newblock Chinese named entity recognition with character-word mixed embedding.
\newblock In \emph{Proceedings of the 2017 ACM on Conference on Information and
  Knowledge Management}, pages 2055--2058. ACM.

\bibitem[{Gong et~al.(2019)Gong, Tang, Zhou, Hao, and Wang}]{gong2019chinese}
Cheng Gong, Jiuyang Tang, Shengwei Zhou, Zepeng Hao, and Jun Wang. 2019.
\newblock Chinese named entity recognition with bert.
\newblock \emph{DEStech Transactions on Computer Science and Engineering},
  (cisnrc).

\bibitem[{Gui et~al.(2019)Gui, Zou, Zhang, Peng, Fu, Wei, and
  Huang}]{gui2019lexicon}
Tao Gui, Yicheng Zou, Qi~Zhang, Minlong Peng, Jinlan Fu, Zhongyu Wei, and
  Xuanjing Huang. 2019.
\newblock A lexicon-based graph neural network for chinese ner.
\newblock In \emph{Proceedings of the 2019 Conference on Empirical Methods in
  Natural Language Processing and the 9th International Joint Conference on
  Natural Language Processing (EMNLP-IJCNLP)}, pages 1039--1049.

\bibitem[{Guo et~al.(2004)Guo, Jiang, Hu, and Zhang}]{guo2004chinese}
Honglei Guo, Jianmin Jiang, Gang Hu, and Tong Zhang. 2004.
\newblock Chinese named entity recognition based on multilevel linguistic
  features.
\newblock In \emph{International Conference on Natural Language Processing},
  pages 90--99. Springer.

\bibitem[{He and Sun(2017{\natexlab{a}})}]{he2017f}
Hangfeng He and Xu~Sun. 2017{\natexlab{a}}.
\newblock F-score driven max margin neural network for named entity recognition
  in chinese social media.
\newblock In \emph{Proceedings of the 15th Conference of the European Chapter
  of the Association for Computational Linguistics: Volume 2, Short Papers},
  pages 713--718.

\bibitem[{He and Sun(2017{\natexlab{b}})}]{he2017unified}
Hangfeng He and Xu~Sun. 2017{\natexlab{b}}.
\newblock A unified model for cross-domain and semi-supervised named entity
  recognition in chinese social media.
\newblock In \emph{Thirty-First AAAI Conference on Artificial Intelligence}.

\bibitem[{Ji and Grishman(2005)}]{ji2005improving}
Heng Ji and Ralph Grishman. 2005.
\newblock Improving name tagging by reference resolution and relation
  detection.
\newblock In \emph{Proceedings of the 43rd Annual Meeting on Association for
  Computational Linguistics}, pages 411--418.

\bibitem[{Johnson et~al.(2020)Johnson, Shen, and Liu}]{johnson2020cwpcbiatt}
Shardrom Johnson, Sherlock Shen, and Yuanchen Liu. 2020.
\newblock Cwpc\_biatt: Character--word--position combined bilstm-attention for
  chinese named entity recognition.
\newblock \emph{Information}, 11(1):45.

\bibitem[{Lample et~al.(2016)Lample, Ballesteros, Subramanian, Kawakami, and
  Dyer}]{lample2016neural}
Guillaume Lample, Miguel Ballesteros, Sandeep Subramanian, Kazuya Kawakami, and
  Chris Dyer. 2016.
\newblock Neural architectures for named entity recognition.
\newblock In \emph{Proceedings of NAACL-HLT}, pages 260--270.

\bibitem[{Levow(2006)}]{levow2006third}
Gina-Anne Levow. 2006.
\newblock The third international chinese language processing bakeoff: Word
  segmentation and named entity recognition.
\newblock In \emph{Proceedings of the Fifth SIGHAN Workshop on Chinese Language
  Processing}, pages 108--117.

\bibitem[{Li et~al.(2014)Li, Hagiwara, Li, and Ji}]{li2014comparison}
Haibo Li, Masato Hagiwara, Qi~Li, and Heng Ji. 2014.
\newblock Comparison of the impact of word segmentation on name tagging for
  chinese and japanese.
\newblock In \emph{LREC}, pages 2532--2536.

\bibitem[{Li et~al.(2020)Li, Sun, Han, and Li}]{li2020survey}
Jing Li, Aixin Sun, Jianglei Han, and Chenliang Li. 2020.
\newblock A survey on deep learning for named entity recognition.
\newblock \emph{IEEE Transactions on Knowledge and Data Engineering}.

\bibitem[{Liu et~al.(2019)Liu, Xu, Xu, Song, and Zu}]{liu2019encoding}
Wei Liu, Tongge Xu, Qinghua Xu, Jiayu Song, and Yueran Zu. 2019.
\newblock \href {https://www.aclweb.org/anthology/N19-1247} {An encoding
  strategy based word-character {LSTM} for {C}hinese {NER}}.
\newblock In \emph{Proceedings of the 2019 Conference of the North {A}merican
  Chapter of the Association for Computational Linguistics: Human Language
  Technologies}, pages 2379--2389.

\bibitem[{Liu et~al.(2010)Liu, Zhu, and Zhao}]{liu2010chinese}
Zhangxun Liu, Conghui Zhu, and Tiejun Zhao. 2010.
\newblock Chinese named entity recognition with a sequence labeling approach:
  based on characters, or based on words?
\newblock In \emph{International Conference on Intelligent Computing}, pages
  634--640. Springer.

\bibitem[{Lu et~al.(2016)Lu, Zhang, and Ji}]{lu2016multi}
Yanan Lu, Yue Zhang, and Dong-Hong Ji. 2016.
\newblock Multi-prototype chinese character embedding.
\newblock In \emph{LREC}.

\bibitem[{Luo and Yang(2016)}]{luo2016empirical}
Wencan Luo and Fan Yang. 2016.
\newblock An empirical study of automatic chinese word segmentation for spoken
  language understanding and named entity recognition.
\newblock In \emph{Proceedings of the 2016 Conference of the North American
  Chapter of the Association for Computational Linguistics: Human Language
  Technologies}, pages 238--248.

\bibitem[{Ma and Hovy(2016)}]{ma2016end}
Xuezhe Ma and Eduard Hovy. 2016.
\newblock End-to-end sequence labeling via bi-directional lstm-cnns-crf.
\newblock In \emph{Proceedings of the 54th Annual Meeting of the Association
  for Computational Linguistics (Volume 1: Long Papers)}, volume~1, pages
  1064--1074.

\bibitem[{Mao et~al.(2008)Mao, Dong, He, Bao, and Wang}]{mao2008chinese}
Xinnian Mao, Yuan Dong, Saike He, Sencheng Bao, and Haila Wang. 2008.
\newblock Chinese word segmentation and named entity recognition based on
  conditional random fields.
\newblock In \emph{Proceedings of the Sixth SIGHAN Workshop on Chinese Language
  Processing}.

\bibitem[{Peng and Dredze(2015)}]{peng2015named}
Nanyun Peng and Mark Dredze. 2015.
\newblock Named entity recognition for chinese social media with jointly
  trained embeddings.
\newblock In \emph{Proceedings of the 2015 Conference on Empirical Methods in
  Natural Language Processing}, pages 548--554.

\bibitem[{Peng and Dredze(2016)}]{peng2016improving}
Nanyun Peng and Mark Dredze. 2016.
\newblock Improving named entity recognition for chinese social media with word
  segmentation representation learning.
\newblock In \emph{The 54th Annual Meeting of the Association for Computational
  Linguistics}, page 149.

\bibitem[{Peters et~al.(2018)Peters, Neumann, Iyyer, Gardner, Clark, Lee, and
  Zettlemoyer}]{peters2018deep}
Matthew Peters, Mark Neumann, Mohit Iyyer, Matt Gardner, Christopher Clark,
  Kenton Lee, and Luke Zettlemoyer. 2018.
\newblock Deep contextualized word representations.
\newblock In \emph{Proceedings of the 2018 Conference of the North American
  Chapter of the Association for Computational Linguistics: Human Language
  Technologies, Volume 1 (Long Papers)}, pages 2227--2237.

\bibitem[{Ratinov and Roth(2009)}]{Ratinov2009CoNLL}
Lev Ratinov and Dan Roth. 2009.
\newblock Design challenges and misconceptions in named entity recognition.
\newblock In \emph{Proceedings of the Thirteenth Conference on Computational
  Natural Language Learning}, pages 147--155.

\bibitem[{Xu et~al.(2013)Xu, Wang, Liu, Liu, Fan, Qian, Tsujii, and
  Chang}]{xu2013joint}
Yan Xu, Yining Wang, Tianren Liu, Jiahua Liu, Yubo Fan, Yi~Qian, Junichi
  Tsujii, and Eric~I Chang. 2013.
\newblock Joint segmentation and named entity recognition using dual
  decomposition in chinese discharge summaries.
\newblock \emph{Journal of the American Medical Informatics Association},
  21(e1):e84--e92.

\bibitem[{Yadav and Bethard(2019)}]{yadav2019survey}
Vikas Yadav and Steven Bethard. 2019.
\newblock A survey on recent advances in named entity recognition from deep
  learning models.
\newblock \emph{arXiv preprint arXiv:1910.11470}.

\bibitem[{Yang et~al.(2018)Yang, Liang, and Zhang}]{yang2018design}
Jie Yang, Shuailong Liang, and Yue Zhang. 2018.
\newblock Design challenges and misconceptions in neural sequence labeling.
\newblock In \emph{Proceedings of the 27th International Conference on
  Computational Linguistics}, pages 3879--3889.

\bibitem[{Zhang and Yang(2018)}]{zhang2018chinese}
Yue Zhang and Jie Yang. 2018.
\newblock Chinese ner using lattice lstm.
\newblock In \emph{Proceedings of the 56th Annual Meeting of the Association
  for Computational Linguistics (Volume 1: Long Papers)}, pages 1554--1564.

\bibitem[{{Zhao} et~al.(2019){Zhao}, {Xu}, and {Cao}}]{8855350}
H.~{Zhao}, M.~{Xu}, and J.~{Cao}. 2019.
\newblock Pre-trained language model transfer on chinese named entity
  recognition.
\newblock In \emph{2019 IEEE 21st International Conference on High Performance
  Computing and Communications; IEEE 17th International Conference on Smart
  City; IEEE 5th International Conference on Data Science and Systems
  (HPCC/SmartCity/DSS)}, pages 2150--2155.

\bibitem[{Zhu and Wang(2019)}]{zhu2019canner}
Yuying Zhu and Guoxin Wang. 2019.
\newblock Can-ner: Convolutional attention network for chinese named entity
  recognition.
\newblock In \emph{Proceedings of the 2019 Conference of the North American
  Chapter of the Association for Computational Linguistics: Human Language
  Technologies, Volume 1 (Long and Short Papers)}, pages 3384--3393.

\end{thebibliography}
\bibliographystyle{acl_natbib}

\end{document}